\pdfoutput=1

\documentclass[11pt]{article}

\usepackage{ACL2023}

\usepackage{times}
\usepackage{latexsym}
\usepackage{gb4e}
\usepackage{amsmath}
\usepackage{bm}
\usepackage{paralist}
\noautomath

\usepackage{setspace}
\usepackage{tcolorbox}
\usepackage{soul}
\setstcolor{red}
\usepackage{todonotes}
{\begin{tcolorbox}[colframe=red!75!black]}%
  {\end{tcolorbox}}

\usepackage[T1]{fontenc}

\usepackage[utf8]{inputenc}

\usepackage{csquotes}

\usepackage{microtype}

\usepackage{inconsolata}

\usepackage{enumitem}
\usepackage{booktabs}
\usepackage{multirow}
\usepackage{xspace}
\usepackage[normalem]{ulem}
\useunder{\uline}{\ul}{}

\newcommand{\entails}{\qquad $\models$ \qquad}

\newcommand{\entlmt}{\textsc{entailment}\xspace}
\newcommand{\ntrl}{\textsc{neutral}\xspace}
\newcommand{\cntr}{\textsc{contradiction}\xspace}

\title{Adverbs, Surprisingly}

\begin{document}

\author{Dmitry Nikolaev$^1$ \quad Collin F. Baker$^2$ \quad Miriam R.L. Petruck$^3$ \quad Sebastian Pad\'{o}$^1$ \\
          $^1$ IMS, University of Stuttgart, Germany \\
          $^2$ International Computer Science Institute, Berkeley, USA \\
          $^3$ FrameNet \\
          Corresponding Author: \texttt{dnikolaev@fastmail.com}}
 
\maketitle

\begin{abstract}
This paper begins with the premise
that 
adverbs are 
neglected in computational linguistics. 
This view derives from two analyses: a literature review 
and a novel adverb dataset to probe a state-of-the-art language model, thereby uncovering systematic gaps
in accounts for adverb meaning. 
We suggest that using Frame Semantics 
for characterizing word meaning, as 
in FrameNet, provides a promising approach to adverb analysis, 
given its ability to describe ambiguity, semantic roles, and null instantiation. 
\end{abstract}

\section{Introduction}
\label{sec:introduction}


Adverbs are 
the part of speech (POS) that has seen the least
attention in (computational) linguistics, likely due to its challenging
nature \citep{conlon-evens-1992-computers}. As \citet[563]{huddleston_pullum_2002} state,
\enquote{the adverb is a [\dots] residual category [\dots] to which words are assigned if they do not satisfy the more specific criteria for nouns, verbs, adjectives, prepositions, and conjunctions.} 

Syntactically, they 
modify many POSs, except nouns
(\textit{eat porridge quickly, hardly noticeable}), or even complete
clauses (\textit{Probably, I'll come tomorrow}). They are 
semantically varied \citep{thomason-stalnaker-73}, ranging from
intensifiers/modifiers (\textit{absolutely}, \textit{beautifully}) to temporal
and spatial specifications (\textit{yesterday, forward}), to
so-called \textit{speaker-oriented adverbs} yielding inferences
about 
speaker 
attitudes, beliefs, and evaluations.
Finally, adverbs can occupy different positions in sentences, creating complex issues of
scoping and ambiguity \citep{ALEXIADOU2004677,d06a4ea3203c4175ba571fa2c55cb6f9}. Consider the following sentences:\footnote{\citet[575]{huddleston_pullum_2002}}

\begin{exe}
\singlespacing
\ex \label{ex:happily}
\begin{xlist}[\hspace{-1em}]
\ex \uline{Happily}, they watched TV until dinner.\label{ex:happilyfront}
\ex They \uline{happily} watched TV until dinner.\label{ex:happilycenter}
\ex They watched TV \uline{happily} until dinner.
\ex They watched TV until dinner \uline{happily}.\label{ex:happilyend}
 \end{xlist}
\end{exe}

\noindent
While 
language users tend to interpret 
Ex.~\ref{ex:happilycenter}--\ref{ex:happilyend}
as describing the TV watchers'
mental state, 
Ex.~\ref{ex:happilyfront}
is ambiguous and can also be read as a positive evaluation of the situation by the speaker.




In sum, adverbs provide crucial information not just about the where and how of events, but also about attitudes and evaluations. However, 
relatively little research on adverbs exists in computational linguistics, although lexical factors are generally recognized
as central for many NLP tasks \citep{10.1145/345508.345576}. Lexical information is generally represented either in online dictionaries or by
embeddings extracted from corpora \citep{Turney2010,devlin-etal-2019-bert,peters-etal-2018-deep}. 
As a dictionary, WordNet \citep{george90:_five_wordn} lists adverbs
but only provides a relatively impoverished account, while lexicons
for sentiment analysis
\citep{Benamara2007SentimentAA,dragut-fellbaum-2014-role} and hedging detection \citep{jeon-choe-2009-key,islam-etal-2020-lexicon} only
consider specific subtypes of adverbs 
as to how they modulate the 
intensity of adjectives. 
On the distributional side, adverbs have been considered
from a derviational perspective
\cite{lazaridou-etal-2013-compositional}; yet, they are rarely scrutinized in
detail.  Among the standard benchmarks, only GLUE
\citep{wang-etal-2018-glue} and BLiMP \citep{warstadt2020blimp}
cover adverbs, and then only marginally. The same is true
of approaches that combine dictionaries and embeddings
\citep{faruqui-etal-2015-retrofitting}. As a consequence, SOTA
language models consistently struggle with adverb meaning, as Section \ref{sec:treatment}  will demonstrate empirically.

This paper argues that Frame Semantics
\citep{fillmore85:_frames_seman_under}, as realized in
FrameNet (FN) \citep{Ruppenhoferetal16}, provides an efficacious framework to
articulate the relevant aspects of adverb meaning. Specifically, as Ex. \ref{ex:happily} illustrates, lexical ambiguity
is captured in terms of frame ambiguity. Moreover, inferences about the 
arguments of adverbs, typically filled by the speaker and the 
lexical unit that the adverb modifies, can be captured and characterized 
via the frame elements (i.e.\ semantic roles) of the frame. 
Notably, FrameNet mechanisms will account for
null-instantiated roles, allowing it to
hint at unexpressed content in cases like
Example~\ref{ex:speakerNI} (v. Section~\ref{sec:appr-fram-analys} for details).

\begin{exe}
  \ex \label{ex:reported} 
  \begin{xlist}
  \ex{} [$_{\textsc{Speaker}}$ The Minister] \textbf{reported} [$_{\textsc{Message}}$ that the cost had exploded].\label{ex:speaker}
  \ex{} [$_{\textsc{Message}}$ The cost had] \textbf{reportedly} [$_{\textsc{Message}}$ exploded].\label{ex:speakerNI}
  \end{xlist} 
\end{exe}

\noindent
In such cases specifically, FrameNet considerations of frame element realization help
to explain the absence of the \textsc{Speaker} semantic role in \ref{ex:speakerNI}.

\paragraph{Plan of the Paper.}
Section~\ref{sec:motivation-nlp} defines the scope of this paper %
(speaker-oriented adverbs) and shows the lack of accounts for
adverbs in NLP through a literature review. Section~\ref{sec:demo}
presents a probing dataset for speaker-oriented adverbs on the basis of
which it demonstrates empirically that current large language models do
not provide accounts for adverb meaning.
%
Section~\ref{sec:adverbs-in-framenet} provides 
general background information on FrameNet, gives details on 
the framework's approach to the description of adverb meaning, and suggests its 
use to improve NLP models.
Section~\ref{sec:discussion} concludes the paper.

\section{Scope and Motivation}
\label{sec:motivation-nlp}

\subsection{Scope}
Given the variety and heterogeneity of adverbs, we restrict the empirical scope of this paper to a subclass of them -- even though we believe that the conceptual points apply to adverbs generally. We focus on
\textit{speaker-oriented adverbs} \citep{Ernst:2009aa}. This 
broad class of adverbs, itself comprises several subtypes
brought together by their
giving rise to a range of inferences
about attitudes and beliefs of the speaker, such as epistemic
beliefs (Ex.~\ref{ex:epistemic}), evaluations
(Ex.~\ref{ex:happily} and~\ref{ex:eval}), and speech acts
(Ex.~\ref{ex:speech-acts}):
\begin{exe}
  \ex \label{ex:epistemic} Peter says: ``Paul is \textbf{certainly} right''. \entails Peter is certain that Paul is right.
  \ex \label{ex:eval} Peter says: ``\textbf{Unfortunately}, Paul arrived''. \entails
  Peter is unhappy that Paul arrived.
  \ex \label{ex:speech-acts} Peter says: ``\textbf{Frankly}, Paul annoys me.'' \entails
  Peter voices his frank opinion.
\end{exe}
Structurally, these entailments are similar to entailments that arise from implicative verbs \cite{10.2307/412084}. As sources of information about how speakers assess states of affairs, they are highly relevant for tasks like opinion mining \cite{Pang+Lee:08b} and
stance detection \citep{thomas-etal-2006-get}. However, while implicative verbs have received considerable attention in the context
of textual entailment~\cite{karttunen-2012-simple,lotan-etal-2013-truthteller}, speaker-oriented adverbs have not.

\subsection{Treatment of Adverbs in Computational Linguistics}
\label{sec:treatment}

This section summarizes work on adverbs in computational linguistics
in the four most relevant areas: WordNets, applications,
distributional modeling, and semantic annotation. Section
\ref{sec:demo} covers large language models separately.

\paragraph{WordNets.} 
Princeton WordNet (WN, version 1.3) \citep{george90:_five_wordn}  covers about 4,500 English adverbs, 
comprising both single words and adverbial multi-word expressions like 
\textit{a priori}. The information recorded includes senses (although most 
adverbs are monosemous) and semantic relations: almost all single-word adverbs are linked to the adjectives from which they are derived,
and some adverbs have antonyms. 
However, WN has no information on the adverbs' syntactic or semantic behavior. The approach of corresponding WordNet resources varies substantially: GermaNet, for German, does not treat adverbs at all \citep{hamp-feldweg-1997-germanet}. In contrast, plWordNet \citep{maziarz-etal-2016-adverbs} provides
a considerably richer description of adverbs, notably regarding lexical relations, but is only available for Polish.

\paragraph{NLP applications.} Apparently, sentiment and emotion analysis 
are the NLP applications that have paid the most attention to adverbs \citep{Benamara2007SentimentAA,dragut-fellbaum-2014-role,chauhan20}. 
Hedge detection, that is, the recognition of
expressions that modulate speaker
confidence in their statements boasts additional work on adverbs \citep{jeon-choe-2009-key,islam-etal-2020-lexicon}.
However, these studies, are generally limited to two specific subtypes: 
scalar adverbs that modify sentiment strength (intensifiers/minimizers: \textit{very/hardly nice}) and adverbs that modify confidence  (\textit{certainly/apparently}). \citet{haider21} also considers locative and temporal adverbs. Confidence-modifying adverbs form a subtype of the speaker-oriented adverbs addressed here, but existing studies 
do not offer a general account of 
these adverbs beyond the requirements of specific tasks.

Studies on structured sentiment and emotion analysis
\citep{barnes-etal-2021-structured,kim-klinger-2018-feels} assume a
different perspective. These works concentrate on defining and
modeling the relations between sentiment- and emotion- introducing
expressions and their semantic arguments, such as the experiencer of
the affect and its target. As the comparison with
Example~\ref{ex:reported} shows, these relations are at times tied
to adverb meanings. However, we are not aware of studies
in this area that deal specifically with adverbs.

\paragraph{Distributional modeling.} A number of studies investigated
the interplay between word embeddings and morphology, analyzing
similarity by parts of speech
\citep{cotterell-schutze-2015-morphological} or investigating
meaning shifts corresponding to morphological derivation
\cite{lazaridou-etal-2013-compositional,pado-etal-2016-predictability}.
Typically, these studies include adverbs, and not surprisingly find
that adverbs behave highly inconsistently.

\paragraph{Semantic annotation.} In principle, frameworks for the annotation of (semantic) argument structure are promising sources for information about adverb meaning, but they
differ widely in the information that they offer.
The PropBank \citep{palmer2005proposition} annotation scheme offers a
range of modifier roles (ARGM) for the annotation of modifiers,
including adverbs. However, the most fitting of these roles, ARGM-ADV,
is a \enquote{catch-all} category. In addition, the PropBank analysis does not
treat adverbs as predicates in their own right and does not assign
roles to them. Thus, \textit{\textbf{fortunately}, she accepted} and
\textit{\textbf{even} she accepted} would receive the same analysis.

In contrast, UCCA \citep{abend-rappoport-2013-universal} explicitly splits adverbs into adverbial modifiers proper (D) and ground elements (G), where the latter
expresses the speaker’s attitude toward the event. However, UCCA does not make the structural relations explicit either.

AMR \citep{banarescu2013abstract} offers a more nuanced approach: many 
adverbs are mapped to their underlying predicates and endowed with complete argument structure,\footnote{For example, AMR treats \textit{sing}
in \textit{sing beautifully} as the first argument of \texttt{beautiful-02}.} while others are interpreted as degree, 
manner, or time modifiers. However, no provision exists in the representation for speaker-oriented adverbs. To illustrate, the AMR annotation of \textit{thankfully, she accepted the present} either treats 
the adverb as describing a general state of affairs (\textit{it is good that she accepted}) or simply omits it.

Finally, Frame Semantics \citep{fillmore85:_frames_seman_under} offers the conceptual infrastructure to improve on these treatments and avoid their 
limitations. Section \ref{sec:adverbs-in-framenet} provides justification of this understanding.

\section{Case Study: Modeling Adverb Meaning as Natural Language Inference}
\label{sec:demo}

One possibility, so far not mentioned, is that the knowledge inherent
in large neural language models might provide a sufficient account of
the meaning of (speaker-oriented) adverbs. In that case, at least from
the NLP perspective, no (new) specific treatment would be
required. However, this state of affairs is not the case, as we show below.

\subsection{Creating Probing Datasets}

To operationalize \enquote{a sufficient account,} we ask
language models to distinguish between valid and invalid inferences
along the lines of Examples~\ref{ex:epistemic}--\ref{ex:speech-acts}.
As input data, we constructed probing examples with inferences for
speaker-oriented adverbs.

We examined four classes of adverbs, motivated by current FrameNet
frames containing adverbs (see Section~\ref{sec:curr-stat-adverbs} for
details). These are: likelihood adverbs (e.g.\
\textit{undoubtedly}, \textit{probably}); unattributed-information
adverbs (\textit{reportedly}, \textit{allegedly},
\textit{supposedly}); degree adverbs (\textit{at least},
\textit{approximately}); and obviousness adverbs (\textit{blatantly},
\textit{conspicuously}).  



We built the datasets from combinations of premises and hypotheses
containing such adverbs, formulated as templates with sets of fillers
for the adverbs and different participant positions. In this manner,
we assessed the LM's capabilities irrespective of specific word
choice. We paired each premise with two to four unambiguous
hypotheses depending on the adverb class. The premise either implies
or contradicts the hypothesis. Table~\ref{tab:template} shows an
example. Hypothesis 1 negates the premise and constitutes a
contradiction. Hypothesis 2 is a valid inference about speaker
evaluation; and Hypothesis 3 is a valid inference about the
uncertainty inherent in the premise.  
%

We report studies on two datasets with different emphases. We designed the first
to be \textit{naturalistic}, based on existing
sentences for adverbs in FrameNet.
Given the
limited size of this dataset, we also created a larger
\textit{synthetic} dataset with simpler, 
more varied, sentences.
The Appendix lists full details on both datasets.

\paragraph{Naturalistic Dataset.}
As stated, we created this dataset based on sentences in the
FrameNet database containing adverbs of the four classes enumerated above. We \enquote{templatized} the sentences 
by treating the
position of the adverb as a slot that can be filled by all
semantically congruent adverbs from the respective class. In 
sentences where the subject is a personal name, we also treated the
subject position as a slot, which we filled with twenty female and
male names popular in the United States.  Because the low number of
sentences of the each type in the FrameNet database, 
and most templates have only one slot, viz.\ the adverb, the
size of this dataset is limited. See Table~\ref{tab:aggregate}
for example counts by adverb class.

\begin{table}[tb]
    \centering
    \small
    \begin{tabular}{lp{5.5cm}}
    \toprule 
        Premise & The celebration had 
been postponed, \textbf{ostensibly} because of the Gulf War  \\ \midrule
 Hyp 1         & The Gulf War \textbf{ostensibly} had no effect on the celebration (\cntr) \\
         Hyp 2 & Someone said that the celebration was postponed because of the Gulf War (\entlmt) \\ 
Hyp 3 & The Gulf War may have had no effect on the celebration (\entlmt)         \\
\bottomrule
    \end{tabular}
    \caption{Naturalistic dataset: Probing items}
    \label{tab:template}
\end{table}

\paragraph{Synthetic Dataset.}
The goal of this dataset was to test if the performance of the
model is robust with regard to the replacement of the main-event
description and varying syntactic complexity of the premises and
hypotheses. It covers three of the four adverb classes:
unattributed-information, degree, and obviousness, where the templates
from the first dataset were most restricted. In these templates,
subjects are always exchangeable. In addition, we also varied the
description of the main action or relation described the
sentence. 

Table~\ref{tab:template-synthetic} shows the template set for
unattributed-information adverbs. The set of adverbs for this class
comprises \textit{reportedly}, \textit{allegedly},
\textit{supposedly}, \textit{apparently}, and
\textit{ostensibly}. Fillers of the \textsc{action} slot include both
gerund phrases (e.g.\ \textit{selling the house}) and noun phrases
(e.g.\ \textit{the wedding}). Entailments and contradictions are
produced in pairs. For entailments, we test two valid inferences
triggered by the adverb. For contradictions, we test embedded clauses
with and without negation.  Table~\ref{tab:synthetic-data-results}
shows the example count for each input type.

\begin{table}[tb]
    \centering
    \small
    \begin{tabular}{lp{5.5cm}}
    \toprule 
    Premise & SUBJ1 said that SUBJ2 \textbf{ADV} opposed ACTION \\ \midrule
    Hyp 1 & SUBJ1 said that SUBJ2 may have opposed ACTION (\entlmt) \\
    Hyp 2 &  SUBJ1 is not sure that SUBJ2 opposed ACTION (\entlmt) \\
    Hyp 3 & SUBJ1 is sure that SUBJ2 opposed ACTION (\cntr) \\
    Hyp 4 & SUBJ1 is sure that SUBJ2 did not support ACTION (\cntr) \\
    \bottomrule
    \end{tabular}
    \caption{Synthetic dataset: Probing items}
    \label{tab:template-synthetic}
\end{table}



\subsection{Probing Setup: NLI models}


Arguably the best match for these types of datasets are the family of
language models optimized for the task of natural-language inference
\citep{storks2019nli}. Concretely, we evaluated the series of NLI
models released by \citet{nie-etal-2020-adversarial}, the SNLI or
Stanford Natural Language Inference models. These models carry out a
three-way classification between \entlmt, \cntr, and \ntrl. The author
fine-tuned their models on a data set created in an iterative,
adversarial, human-in-the-loop fashion, designed to remedy the
shortcomings of previous NLI datasets \citep{belinkov-etal-2019-dont}.
Preliminary experiments with different available base architectures 
(RoBERTa, ALBERT, BART, ELECTRA, and XLNet) showed that
RoBERTa-large\footnote{ynie/roberta-large-snli\_mnli\_fever\_anli\_R1\_R2\_R3-nli}
was the best-performing variant. Thus, we used this model for
evaluations. We used our probing datasets solely for evaluation, not for
further fine-tuning.

For analysis, we checked the labels that the model predicted 
with their corresponding probabilities. In several cases, we performed
additional tests to verify whether the adverbs or other properties of
the sentence determined the model predictions.

\subsection{Evaluation on a Naturalistic Dataset}

\subsubsection{Overall results}

Table~\ref{tab:aggregate} shows overall results of the SNLI model on
the naturalistic dataset for the four adverb classes. The adverb
classes are not strictly comparable because they are represented by
different input sentences (as described above), which include all
types of lexical and syntactic confounds. Nevertheless, our
experiments showed two consistent results: (i)~the model cannot
correctly draw inferences based on some
set of adverbs on which it fails;  
(ii)~the presence of
adverbs increases the difficulty for the model to draw correct
inferences in general.  What follows is a survey of the evidence for these two claims.

\begin{table}[t]
\centering
\small
\begin{tabular}{@{}lll@{}}
\toprule
Adverb class & Error rate (\%) & \# sentences \\ \midrule
Likelihood & 2 & 5,880 \\
\begin{tabular}[c]{@{}l@{}}Unattributed\\ information\end{tabular} & 6 & 90 \\
Degree & 25 & 35 \\
Obviousness & 23 & 16 \\ \bottomrule
\end{tabular}
\caption{Naturalistic dataset: SNLI model error rates by adverb class}
\label{tab:aggregate}
\end{table}



\subsubsection{Failure to Understand Adverbs}

\paragraph{Degree adverbs.} The model does not understand that \textit{at least as big}
is incompatible with \textit{smaller}. While it correctly labels the
pair \textit{Lantau covers nearly twice the area of Hong Kong Island}
-- \textit{Lantau is at least as big as Hong Kong Island} as \entlmt\
and the same premise with \textit{Lantau is much smaller than Hong
Kong Island} as \cntr, it considers that this premise also entails
\textit{Hong Kong Island is at least as big as Lantau}, which is also
a straightforward contradiction.

The quantifier--adverb combination \textit{almost
every} constitutes another weak point of the model. While it correctly labels the pair
\textit{\textbf{Almost all} assignments are challenging in different ways} vs.\ 
\textit{Most of the assignments are difficult}, it labels
\textit{\textbf{Almost every} assignment is a challenge in a different way} vs.\ 
the same as \ntrl.\footnote{The model answers correctly only when there is a larger lexical overlap, as in \textit{Most of the assignments are challenging.}}

\paragraph{Unattributed-information adverbs.} The correct analysis of these adverbs 
is subtle since valid inferences may be expressed in ways that differ from the 
premise both lexically and syntactically.

Sometimes the model answers incorrectly 
with extremely high confidence. The example from Table~\ref{tab:template} is a case in point. \textit{The Gulf War ostensibly
had no effect on the celebration} 
is always correctly labeled as \cntr. The \textit{Someone said...} hypothesis is also correctly labelled as \entlmt\ with \textbf{any} adverb in the premise. Strikingly, the model gives the same result when the adverb is omitted. This  suggests that the model does not take the adverb in the premise into account.

The experiments with Hypothesis 3 (cf. Table~\ref{tab:template})
corroborated that understanding: regardless of the combination of the
adverb in the premise and the hypothesis, the model confidently marks
the pair as \cntr\ or \ntrl\ with almost zero probability attached to
the prediction of \entlmt.  This finding shows that while the model
may be able to draw a positive inference from the hearsay adverb (the
reported event may have happened), it is completely unaware of the
possibility of the negative inference, i.e.\ that the reported event
may not have taken place: 12 times out of 16, the model confidently
predicts \cntr.

\begin{table*}[tb!]
\centering
{\small
\begin{tabular}{@{}cclrrrrr@{}}
\toprule
\textbf{Verb} &
  \textbf{Prediction} &
  \multicolumn{1}{c}{\textbf{Hypothesis}} &
  \multicolumn{1}{c}{\textbf{obviously}} &
  \multicolumn{1}{c}{\textbf{clearly}} &
  \multicolumn{1}{c}{\textbf{publicly}} &
  \multicolumn{1}{c}{\textbf{blatantly}} &
  \multicolumn{1}{c}{\textbf{no adverb}} \\ \midrule
 &
   &
  \cellcolor[HTML]{F3F3F3}Simple &
  \cellcolor[HTML]{F3F3F3}0.94 &
  \cellcolor[HTML]{F3F3F3}0.94 &
  \cellcolor[HTML]{F3F3F3}0.95 &
  \cellcolor[HTML]{F3F3F3}0.96 &
  \cellcolor[HTML]{F3F3F3}0.97 \\
 &
  \multirow{-2}{*}{Entailment} &
  \cellcolor[HTML]{FFFFFF}Complex &
  \cellcolor[HTML]{FFFFFF}0.60 &
  \cellcolor[HTML]{FFFFFF}0.62 &
  \cellcolor[HTML]{FFFFFF}0.70 &
  \cellcolor[HTML]{FFFFFF}0.71 &
  \cellcolor[HTML]{FFFFFF}0.85 \\
 &
   &
  \cellcolor[HTML]{F3F3F3}Simple &
  \cellcolor[HTML]{F3F3F3}0.05 &
  \cellcolor[HTML]{F3F3F3}0.05 &
  \cellcolor[HTML]{F3F3F3}0.05 &
  \cellcolor[HTML]{F3F3F3}0.04 &
  \cellcolor[HTML]{F3F3F3}0.02 \\
\multirow{-4}{*}{\textit{aid}} &
  \multirow{-2}{*}{Neutral} &
  \cellcolor[HTML]{FFFFFF}Complex &
  \cellcolor[HTML]{FFFFFF}0.39 &
  \cellcolor[HTML]{FFFFFF}0.38 &
  \cellcolor[HTML]{FFFFFF}0.29 &
  \cellcolor[HTML]{FFFFFF}0.27 &
  \cellcolor[HTML]{FFFFFF}0.15 \\ \midrule
 &
   &
  \cellcolor[HTML]{F3F3F3}Simple &
  \cellcolor[HTML]{F3F3F3}0.92 &
  \cellcolor[HTML]{F3F3F3}0.92 &
  \cellcolor[HTML]{F3F3F3}0.92 &
  \cellcolor[HTML]{F3F3F3}0.95 &
  \cellcolor[HTML]{F3F3F3}0.97 \\
 &
  \multirow{-2}{*}{Entailment} &
  \cellcolor[HTML]{FFFFFF}Complex &
  \cellcolor[HTML]{FFFFFF}0.53 &
  \cellcolor[HTML]{FFFFFF}0.52 &
  \cellcolor[HTML]{FFFFFF}0.58 &
  \cellcolor[HTML]{FFFFFF}0.61 &
  \cellcolor[HTML]{FFFFFF}0.77 \\
 &
   &
  \cellcolor[HTML]{F3F3F3}Simple &
  \cellcolor[HTML]{F3F3F3}0.07 &
  \cellcolor[HTML]{F3F3F3}0.08 &
  \cellcolor[HTML]{F3F3F3}0.08 &
  \cellcolor[HTML]{F3F3F3}0.05 &
  \cellcolor[HTML]{F3F3F3}0.03 \\
\multirow{-4}{*}{\textit{help}} &
  \multirow{-2}{*}{Neutral} &
  \cellcolor[HTML]{FFFFFF}Complex &
  \cellcolor[HTML]{FFFFFF}{\ul 0.47} &
  \cellcolor[HTML]{FFFFFF}{\ul 0.47} &
  \cellcolor[HTML]{FFFFFF}{\ul 0.41} &
  \cellcolor[HTML]{FFFFFF}0.38 &
  \cellcolor[HTML]{FFFFFF}0.22 \\ \midrule
 &
   &
  \cellcolor[HTML]{F3F3F3}Simple &
  \cellcolor[HTML]{F3F3F3}0.99 &
  \cellcolor[HTML]{F3F3F3}0.99 &
  \cellcolor[HTML]{F3F3F3}0.99 &
  \cellcolor[HTML]{F3F3F3}0.99 &
  \cellcolor[HTML]{F3F3F3}0.99 \\
 &
  \multirow{-2}{*}{Entailment} &
  \cellcolor[HTML]{FFFFFF}Complex &
  \cellcolor[HTML]{FFFFFF}0.41 &
  \cellcolor[HTML]{FFFFFF}0.43 &
  \cellcolor[HTML]{FFFFFF}0.57 &
  \cellcolor[HTML]{FFFFFF}0.39 &
  \cellcolor[HTML]{FFFFFF}0.85 \\
 &
   &
  \cellcolor[HTML]{F3F3F3}Simple &
  \cellcolor[HTML]{F3F3F3}0.01 &
  \cellcolor[HTML]{F3F3F3}0.01 &
  \cellcolor[HTML]{F3F3F3}0.01 &
  \cellcolor[HTML]{F3F3F3}0.01 &
  \cellcolor[HTML]{F3F3F3}0 \\
\multirow{-4}{*}{\textit{support}} &
  \multirow{-2}{*}{Neutral} &
  \cellcolor[HTML]{FFFFFF}Complex &
  \cellcolor[HTML]{FFFFFF}\textbf{0.55} &
  \cellcolor[HTML]{FFFFFF}\textbf{0.53} &
  \cellcolor[HTML]{FFFFFF}0.40 &
  \cellcolor[HTML]{FFFFFF}\textbf{0.40} &
  \cellcolor[HTML]{FFFFFF}0.15 \\ \bottomrule
\end{tabular}
}
\caption{Prediction of NLI model given \textit{Castro ADV backed the rebels} as
premise and \textit{Castro VERBed the rebels} or \textit{Castro tried to VERB the rebels}
as hypothesis (\textit{simple} and \textit{complex} respectively).
Boldface indicates wrong model predictions;
underline indicates \enquote{borderline correct} cases where an incorrect label received a probability~$>$ 40\%.}
\label{tab:rebels}
\end{table*}

\subsubsection{Adverbs Complicate Inference}

In another analysis, we investigate the impact of the sentences' structural complexity on prediction quality. We frequently found that the model 
correctly inferred when the hypothesis is structurally simple or no adverb is given, but failed when the hypothesis had an embedded clause and the premise had an adverb. Table~\ref{tab:rebels} shows a concrete example, which permits three observations:
\begin{compactenum}
    \item The model is sensitive to whether the hypothesis contains an embedded clause:
    the confidence for the correct prediction drops from  $\approx$1 to $\approx$0.8 for
    all verbs in the no-adverb case.
    \item The presence of the adverb is not noticeable with structurally simple
    hypotheses: the confidence in the correct answer remains $>$0.9.
    \item The combination of an adverb and an embedded clause can
    derail the model~-- paradoxically most so for the verb \textit{support},
    where the model was most confident without an adverb.
\end{compactenum}
Furthermore, note that an adverb can force the model to change its decision even in the presence of a strong lexical cue. Given the hypothesis \textit{The students were obviously drunk}, the model correctly identifies \textit{The 
students abhor/forswore/renounced alcohol} as \cntr. While the model labels \textit{The students abjured 
alcohol} as \entlmt, (perhaps) because of an incorrect analysis of the verb,
when we change the hypothesis to \textit{The students were \textbf{conspicuously} drunk}, the model confidently and correctly labels \textit{The students abjured 
alcohol} as \cntr.

\begin{table*}[t]
\centering
\small

\begin{tabular}{@{}llrrrrr@{}}
\toprule
\textbf{Semantic type} & \textbf{Test} & \multicolumn{1}{l}{\textbf{Entailment}} & \multicolumn{1}{l}{\textbf{Neutral}} & \multicolumn{1}{l}{\textbf{Contradiction}} & \multicolumn{1}{l}{\textbf{Error rate (\%)}} & \multicolumn{1}{l}{\textbf{\# sentences}} \\ \midrule
 & Entailment 1 & \cellcolor[HTML]{F3F3F3}70,188 & 12 & 0 & $\approx$ 0 & 70,200 \\
 & Entailment 2 & \cellcolor[HTML]{F3F3F3}134 & 70,066 & 0 & $\approx$ 100 & 70,200 \\
 & Contradiction 1 & 7,940 & 62,260 & \cellcolor[HTML]{F3F3F3}0 & 100 & 70,200 \\
\multirow{-4}{*}{\begin{tabular}[c]{@{}l@{}}Unattributed\\ information\end{tabular}} & Contradiction 2 & 567 & 69,633 & \cellcolor[HTML]{F3F3F3}0 & 100 & 70,200 \\ \midrule
 & Entailment & \cellcolor[HTML]{F3F3F3}31,200 & 0 & 0 & 0 & 31,200 \\
\multirow{-2}{*}{\begin{tabular}[c]{@{}l@{}}Degree (properties\\ of people)\end{tabular}} & Contradiction & 12,390 & 3,980 & \cellcolor[HTML]{F3F3F3}14,830 & 52 & 31,200 \\ \midrule
 & Entailment & \cellcolor[HTML]{F3F3F3}840 & 0 & 0 & 0 & 840 \\
\multirow{-2}{*}{\begin{tabular}[c]{@{}l@{}}Degree (properties\\ of objects)\end{tabular}} & Contradiction & 547 & 0 & \cellcolor[HTML]{F3F3F3}293 & 65 & 840 \\ \midrule
 & Entailment & \cellcolor[HTML]{F3F3F3}38,400 & 0 & 0 & 0 & 38,400 \\ 
\multirow{-2}{*}{Degree (quantities)} & Contradiction & 0 & 0 & \cellcolor[HTML]{F3F3F3}38,400 & 0 & 38,400 \\ \midrule
 & Entailment 1 & \cellcolor[HTML]{F3F3F3}54,600 & 0 & 0 & 0 & 54,600 \\
 & Entailment 2 & \cellcolor[HTML]{F3F3F3}33,217 & 21,383 & 0 & 39 & 54,600 \\
 & Contradiction 1 & 61 & 0 & \cellcolor[HTML]{F3F3F3}54,539 & $\approx$ 0 & 54,600 \\
\multirow{-4}{*}{Obviousness} & Contradiction 2 & 0 & 1,615 & \cellcolor[HTML]{F3F3F3}52,985 & 3 & 54,600 \\ \bottomrule
\end{tabular}
\caption{Synthetic dataset: Model predictions (cells with correct predictions have gray background) for each template class and error rates.}\label{tab:synthetic-data-results}
\end{table*}

\subsection{Evaluation on a Synthetic Dataset}

The results for the application of same model on the
larger synthetic dataset are shown in
Table~\ref{tab:synthetic-data-results}. They demonstrate that in
general the task of drawing correct inferences from adverbs is very
difficult for the model. Instead, the model tends to consistently
predict the same relation (entailment / neutral / contradiction) for
all sentences for an adverb class. It is able to correctly predict
inference for the quantity degree class (\textit{at least two dozen
people} $\models$ \textit{many people} and $\not\models$
\textit{nobody}). However, even syntactically trivial entailments and
contradictions in other classes lead to systematic failures. E.g.,
while the model can correctly identify the inference \textit{James
said that Mary reportedly opposed the wedding} $\models$ \textit{James
said that Mary may have opposed the wedding}, it fails on the
entailment of the type \textit{James \textbf{is not sure} that Mary
opposed the wedding}.

Similarly, with obviousness adverbs, while the examples of the type
\textit{James blatantly criticized Mary} $\models$ \textit{James
disparaged Mary} are easy for the model, entailments like
\textit{James tried to disparage Mary} leads to near-chance
performance. In the domain of adverb-modulated relations, while the
model seems to do well on entailments (\textit{James is at least twice
as rich as Mary} $\models$ \textit{James's net worth is at least as
big as Mary's}), in fact it does not understand that the relation is
not symmetric and therefore cannot correctly identify contradictions
(\textit{Mary's net worth is at least as big as James's}).


\subsection{Discussion}
\label{sec:discussion-1}

Taken together, these experiments demonstrate systematic shortcomings
in the ability of current large language models to account for adverb
meaning, either glossing over them completely or making rather random
inferences about their meaning.  Arguably, this study only looked at a
specific type of language model and other types of language models
would fare better. However, converging evidence from the literature
exists.

For instance, \citet{nikolaev2023representation} analyzed sentence
transformers, which might be expected to provide the most nuanced
understanding of adverbs. Instead, the study found that the sentences'
main participants (subjects and objects) primarily determine the
semantic similarity of sentence pairs, which is largely independent
of adverbs.  The paper argues that this behavior arises from the
structure of the training data for sentence transformers (online
conversations, duplicate questions on WikiAnswers), where sentence
pairs labelled as semantically similar often have similar sets of main
participants (subjects and objects) and can vary widely in other
respects.

If a similar bias is at play in the NLI models in the present study, creating larger, richer training sets that involve adverbs in a systematic manner is a way forward. However, given the relative scarcity of adverbs and their complex behavior (cf.~Section~1),  it seems unlikely that this effect will emerge naturally by pre-training on ever larger datasets. Instead, the evidence provided here indicates that adverb data must be created intentionally. The following section outlines a proposal to do so.


\section{Describing Adverbs in FrameNet}
\label{sec:adverbs-in-framenet}

This section will provide a brief background to FrameNet (Section~\ref{sec:background-to-fn}),
show how FrameNet can be useful for the analysis of adverbs (Section~\ref{sec:appr-fram-analys}),
survey the data on adverbs contained in the current version of the dataset (Section~\ref{sec:curr-stat-adverbs}), and propose concrete directions for next steps (Section~\ref{sec:future-work}).

\subsection{Background to FrameNet} 
\label{sec:background-to-fn} 
FrameNet (\textit{FN}, \citealt{Ruppenhoferetal16}) is a research and resource-development 
project in corpus-based computational lexicography grounded in the theory of \textit{Frame Semantics} \citep{fillmore85:_frames_seman_under}.

At the heart of the work is the \textit{semantic frame}, a script-like
knowledge structure that facilitates inferencing within and across
events, situations, states-of-affairs, relations, and objects. %
FN defines a semantic frame in terms of its \textit{frame elements}
(\textit{FEs}), or participants (and other concepts) in the scene that
the frame captures; a \textit{lexical unit} (LU) is a pairing of a
lemma and a frame, characterizing that LU in terms of the frame that it evokes. 
FN frames may include more than one POS, and FrameNet does not claim that the
LUs of a frame are synonymous, merely that they are semantically similar in referring to the same situation.
Additionally, FN distinguishes between core FEs and non-core FEs; the former uniquely define a frame and the later identify concepts that characterize events or situations more generally, such as time and place.
To illustrate, Example~\ref{ex:buy} shows annotation for the
verb \textit{BUY}, defined in the \texttt{Commerce\_buy} frame, with
the FEs \textsc{{\textcolor{black}{Buyer}}},
\textsc{{\textcolor{black}{Seller}}},
\textsc{{\textcolor{black}{Goods}}}, and
\textsc{{\textcolor{black}{Money}}}.\footnote{This paper uses the
  following typographical conventions: frame names appear in
  \texttt{typewriter font}; FE names are in \textsc{small caps}; and
  lexical units are in \textbf{BOLD CAPS.}}
\begin{exe}
\singlespacing
\ex \label{ex:buy}
 [\textcolor{black}{Chuck} \textcolor{black}{$_{\textsc{Buyer}}$}] \textbf{\textcolor{black}{BOUGHT}}
[\textcolor{black}{a car} \textcolor{black}{$_{\textsc{Goods}}$}]
[\textcolor{black}{from Jerry} \textcolor{black}{$_{\textsc{Seller}}$}] [\textcolor{black}{for \$2,000} \textcolor{black} {$_{\textsc{Money}}$}] \label{buy}
\end{exe}

FrameNet annotators label approximately 20 sentences for each LU in
each frame; and automatic processes tabulate the results to produce
\textit{valence} descriptions, or \textit{semantic-syntactic
combinatorial possibilities} of each LU. These also include \textit{null-instantiated} core FEs, i.e.\ 
FEs that uniquely define a frame, even when not realized linguistically.
Such valence descriptions
provide information about meaning-form mappings  that are important for natural-language understanding. 
FrameNet data, or semantic parsers built from them, have proven 
useful for tasks such as recognizing  paraphrases \citep{ellsworth-janin-2007-mutaphrase}, drawing
inferences \citep{ben-aharon-etal-2010-generating}, machine translation \citep{zhai-etal-2013-handling}, question answering \citep{khashabi18:_quest}, or paraphrasing
\citep{Wang2018ATI}.

At present, the FrameNet database (Release 1.7) holds 1,224 frames, defined in terms of 10,478 frame-specific FEs, and 13,686 LUs. Of those lexical units, 61\% have  \textit{lexicographic} annotation, i.e.\ 
annotation for one target lemma per sentence.

\subsection{FrameNet for the Analysis of Adverbs} 
\label{sec:appr-fram-analys}

We now outline how the descriptive devices of FrameNet, as outlined in Section~\ref{sec:background-to-fn}, can capture
the relevant facts about adverb meaning and 
address the core challenges of adverb classes, ambiguity, inferences, and null instantation of roles.

\paragraph{Frames.}  Since frame definitions encompass much of the meaning of each LU, many FN frames already offer fine-grained, semantically motivated descriptions of adverb classes. For example, the \texttt{Emotion\_directed} frame captures the semantic similarity of \textit{happy}, \textit{happily}, \textit{happiness}, \textit{sad}, and \textit{sadly} and offers
a starting point for the description of emotion-related adverbs, by exploiting the fact 
that these adverbs evoke the same background knowledge as the corresponding LUs of 
other parts of speech \citep{Ruppenhoferetal16}. 

When a lemma is ambiguous, each sense gets mapped to a different frame; each mapping is a separate lexical unit (LU).  For instance, Example~\ref{ex:happily} in Section \ref{sec:introduction} includes the 
lemma \textit{happily}, which is ambiguous:  In Example \ref{ex:happilyfront}, \textit{happily} is defined in the \texttt{Luck} frame (along with \textit{fortunately} and \textit{luckily}).  The definition of this frame indicates that there is someone, the \textsc{Protagonist}, for whom a particular state of affairs is surprisingly good or bad. But this sentence does not express the \textsc{Protagonist}; this is a case of null instantiation or NI (see below for details).
The other three sentences, Examples~\ref{ex:happilycenter}--\ref{ex:happilyend}, illustrate \textit{happily} in the \texttt{Emotion\_directed} frame. This involves an emotional response of someone, the \textsc{Experiencer}, to a stimulus, the \textsc{Stimulus} FE (here, watching TV), which evokes the emotional response, specifically happiness (recoverable from the definition of the LU \textit{happily}).  In these examples, the \textsc{Experiencer} is explicit, so no inference is required 
(although coreference resolution will be required to resolve the referent of \textit{they}).
 Example~\ref{ex:happily2} shows the annotations of the sentences in the \texttt{Luck} frame 
 (Ex.~\ref{ex:happily2front}) and in the \texttt{Emotion\_directed} frame (Ex.~\ref{ex:happily2center}):
\begin{exe}
\singlespacing
\ex \label{ex:happily2}
\begin{xlist}
\ex{} HAPPILY, [\textcolor{black}{they watched TV until dinner} \textcolor{black}{$_\textsc{State\_of\_affairs}$}] \textcolor{black}{$_\textsc{Protagonist: NI}$}.\label{ex:happily2front}
\ex{} [\textcolor{black}{They} \textcolor{black}{$_\textsc{Experiencer}$}] HAPPILY [\textcolor{black}{watched TV until dinner} \textcolor{black}{$_\textsc{Stimulus}$}].\label{ex:happily2center}
 \end{xlist}
\end{exe}

\paragraph{Frame Elements.}  
In FrameNet, frame elements are associated
with (classes of) inferences \citep{chang-etal-2002-putting}. Such inferences can capture important aspects of adverb meaning, as we have shown in Section~\ref{sec:motivation-nlp}. The valence patterns for the two senses of \textit{happily} shown above lead to different inferences via the two sets of frame elements:
\begin{description}

\item \texttt{Luck}: A \textsc{State\_of\_affairs} is evaluated as good (or bad) [...] for a particular \textsc{Protagonist}. 
\item \texttt{Emotion\_directed:} An \textsc{Experiencer} [feels or experiences] a particular emotional response to a \textsc{Stimulus} or about a \textsc{Topic}.
\end{description}
While such natural language descriptions were traditionally hard to formalize, the recent advances in \enquote{prompting} language models \citep{shin-etal-2020-autoprompt} have reestablished natural language descriptions as sufficient in many conditions (cf.\ also our template-based probing dataset in Section~\ref{sec:demo}).


\begin{table*}[tb]
\centering
{\small
\begin{tabular}{p{1.8cm} p{7cm} p{5.5cm}} 
\toprule
\textbf{Frame name} &
  \textbf{Adverbial lexical units \& example sentence} &
  \textbf{Definition} \\ \midrule

\textbf{\texttt{Unattributed information}} &
\textit{allegedly.adv, ostensibly.adv, purportedly.adv, reportedly.adv, supposedly.adv} \newline
\textbf{Ex.} One person was REPORTEDLY killed\dots & 
A speaker presents a \textsc{Reported fact} as deriving from statements (made directly to them or to others) of third parties. \\
\midrule

\textbf{\texttt{Likelihood}}     & \textit{certainly, likely, probably, possibly} \newline
\textbf{Ex.}  This will LIKELY not be enough to stop\dots 
& 
This frame concerns the likelihood of a \textsc{Hypothetical event} occurring, the only core frame element in the frame. \\
\midrule

\textbf{ \texttt{Obviousness}} & 
\textit{audibly.adv, clearly.adv, evidently.adv, noticeably.adv, obviously.adv, visibly.adv} \newline
\textbf{Ex.} It is CLEARLY desirable to permit the gifted youngsters to flourish. &
 A \textsc{Phenomenon} is portrayed in terms of the \textsc{Degree} of likelihood that it will be perceived and known, given the (usually implicit) \textsc{Evidence}, \textsc{Perceiver}, and \textsc{Circumstances} in which it is considered. \\
\midrule

\textbf{\texttt{Degree}} & \textit{a little (bit).adv,  a lot.adv, absolutely.adv, as hell.adv, far.adv, fully.adv, in part.adv, kind of.adv, so.adv, somewhat.adv, that.adv, totally.adv, very.adv, way.adv}\newline
\textbf{Ex.} I had ABSOLUTELY nothing to say. &
LUs in this frame modify a \textsc{Gradable attribute} and describe intensities at the extreme positions on a scale.  \\
\bottomrule
\end{tabular}}
\caption{FrameNet Frames characterizing Speaker-Oriented Adverbs}\label{tab:adverbframes}
\end{table*}

\paragraph{Null instantiation.} FrameNet annotates information about the conceptually required 
\enquote{core} semantic roles of a frame even if absent from the text. FN distinguishes three types 
of null instantiation, one licensed by a
construction and the others licensed lexically.
FrameNet includes approximately 55,700 NI
labels in its annotations; and roughly one-quarter of these
omissions are licensed constructionally, with the remaining 75\%
licensed lexically \citep{petruck-2019-meaning}. 

This capability of FrameNet is particularly important for adverbs, notably speaker-oriented adverbs. By definition, these adverbs welcome inferences about the speaker, who is typically not realized 
unless the statement is part of reported speech or thought: \textit{The father thought: \enquote{Happily they are all watching TV.}}

Returning to Example~\ref{ex:reported} (above), \ref{ex:speaker} illustrates an instantiated \textsc{Speaker} and \ref{ex:speakerNI} illustrates a \textit{null-instantiated} \textsc{Speaker}, a fact that FN records in its database. 
No other lexical resource used extensively in computational linguistics records such information.

\subsection{Current Status of Adverbs in FrameNet}
\label{sec:curr-stat-adverbs}

Currently, FrameNet (Release 1.7) contains 217 adverb LUs.  Of these adverbs, 158 have annotation, with a total of 2,475 annotations of adverbs on sentences in the database, yielding a mean of 16 annotations per LU.  However, like many linguistic phenomena, the annotations exhibit a highly skewed (Zipfian) distribution: 41 of the 158 LUs  have only one annotation while nine have more than 50 annotations each. In line with its general principles, FrameNet chose not to define one single frame to capture all speaker-oriented adverbs, instead defining each such adverb according to the specific frame it evokes. At the same time, the class of speaker-oriented adverbs is arguably recoverable from the union of a set of frames all of which support inferences about the speaker by way of describing the speaker through a certain frame element. In this way, the existing frames and their annotations provide a suitable basis for creating data for this (and future) research.
    

Table \ref{tab:adverbframes} shows the four FrameNet frames used
to suggest adverbs for the experiment described in Section~\ref{sec:demo} together with the adverbs listed, illustrative example sentences, and their definitions.


\noindent



\subsection{Next Steps}
\label{sec:future-work}

As the numbers show (Section~\ref{sec:curr-stat-adverbs}), 
FrameNet has not attended to adverbs either.
Perhaps this fact represents 
a principal incompatibility: the description of adverbs may 
not welcome using concepts that FN developed for traditional predicates with clear-cut valence.
Yet, we believe that 
including adverbs in FrameNet both follows the spirit of what \citet{fillmore85:_frames_seman_under} called \enquote{semantics of understanding} and is in line with FrameNet practice. Still, it will require work on two principal levels: theoretical development and practical lexicographic analysis.

At the theoretical level, the FrameNet approach has seen constant development over the 25 years of the project's existence. 
In  initial verb-centered frames, nominals tended to fill FEs, 
with additional attributes realized as adverbs. Next,  FN added deverbal nouns to frames, 
which largely take the same frame elements. To expand to other types of nouns, like natural kinds and artifacts, 
FrameNet broadened the concept of FE to encompass \textit{qualia} such as 
substance or purpose \citep{pustejovsky-1991-generative}.
Layering the annotation of nouns as FEs of verbs, and modifiers of nouns as \textit{their} FEs 
provided a richer semantic representation. Next,  FrameNet included adjectives as frame-evoking elements, permitting 
generalizations over domains like speed or temperature. While most aspects of adverbs description are already present in FrameNet (cf. above), theoretical analysis must make precise the implications of annotating null instantiated adverbial frame elements at scale.

At the practical level, 
the time is ripe to add many more adverbs to appropriate existing frames and to create new frames for adverbs as needed.
The principles of annotating naturally occurring 
text and extracting 
valence descriptions for LUs established on the other parts of speech carry over to adverbs. The combination of valence descriptions and  annotated instances
constitute essential inputs to characterize inferences.


%

Clearly, the more annotation, the better, but large-scale expert 
annotation is slow and resource-intensive. Using crowdsourcing, which
permits parallelizing (thus, speeding up) annotation, is a possible
mitigation.  \citet{fossati-etal-2013-outsourcing} and
\citet{feizabadi-pado-2014-crowdsourcing} demonstrated success with
crowdsourcing for frame-semantic annotation when 
the task is narrowed down appropriately. 
Substantial promise exists
to extract adverb annotation automatically from comparable corpora
\citep{roth-frank-2015-inducing} and paraphrasing models
\citep{Wang2018ATI}. Even for the core task of FrameNet analysis,
defining frames, \citet{ustalov-etal-2018-unsupervised-semantic}
proposed automatic methods.
Still, full automation remains hard, given concerns of
quality and consistency.

\section{Conclusion}
\label{sec:discussion}

\citet{conlon-evens-1992-computers} stated 
that adverbs are under-researched in computational linguistics; this statement is still true.
Adverbs have received attention only in 
two applications: 
sentiment analysis and hedging detection. The large language models 
used here 
show systematic gaps 
in capturing 
adverb meaning. 
The problem is \textbf{not} solved.

We propose that Frame Semantics, as
embodied in FrameNet,  along with improved techniques to mitigate the annotation effort to extend FN  with new frames and annotations, can capture the meaning and implicatures of adverbs. Considering frames as lexical constructions
 \citep{fillmore2008border}, this proposal fits well with recent work to combine language models and construction grammar \citep{https://doi.org/10.48550/arxiv.2302.02178}.

Multiple ways exist 
for computational modeling to use such a resource, e.g., by extending the coverage of semantic role labellers to a larger range of adverbs, or by fine-tuning language models on large annotated datasets for which our probing dataset 
can serve as a blueprint.

\section*{Limitations}
We only used English data in the study, so we cannot guarantee that the findings will generalize to other languages (cf.\ \citealt{bender2019rule}). The English NLI datasets are, as usual, larger than for other languages, so we should expect models targeting other languages to have worse performance. We do, however, believe that the challenges of adverbs are comparable in other languages, particularly in typologically similar languages. 


\section*{Ethics Statement}

The paper argues for a new approach to the treatment of adverbs in the development of resources and applications in NLP. We consider better understanding of language by computational models as not posing a significant 
societal risk in itself. The dataset used for the computational experiment in Section~\ref{sec:demo}
was created based on the data contained in the publicly available FrameNet corpus and, as far as we
are aware, does not contain sensitive elements. Implementation of our proposed methodology has the
same risks as any data-driven approach in computational linguistics, but we assume that we cannot
safeguard against its possible misuse due to its very general nature.


\bibliography{anthology,adverbs}
\bibliographystyle{acl_natbib}

\appendix

\section{Details on the Naturalistic Dataset}
\label{sec:appendix-naturalistic}

The probing dataset includes a series of template classes. Each template class
corresponds to an adverb class and contains several NLI templates with slots for
adverbs and, when the structure permits it, also for the subject. In testing, we used
all pairs of adverbs from the relevant class to instantiate the premise
and the hypothesis. When a variable for subject exists in the premise, we used the same subject in the hypotheses.

\subsection{Likelihood Adverbs}

\paragraph{Adverbs:} \textit{undoubtedly}, \textit{surely}, \textit{positively}, \textit{likely}, \textit{certainly}, \textit{definitely}, \textit{totally}.

\paragraph{Fillers for the subject slot:} \textit{Barbara}, \textit{Charles}, \textit{David}, \textit{Elizabeth}, \textit{James}, \textit{Jennifer}, \textit{Jessica}, \textit{John}, \textit{Joseph}, \textit{Karen}, \textit{Linda}, \textit{Mary}, \textit{Michael}, \textit{Patricia}, \textit{Richard}, \textit{Robert}, \textit{Sarah}, \textit{Susan}, \textit{Thomas}, \textit{William}.

\begin{enumerate}
    \item
        \textbf{Premise:} \textit{SUBJ is ADV gonna have to check it tomorrow afternoon again}.\\
        \textbf{Entailment:} \textit{SUBJ is ADV going to have to check it again}.\\
        \textbf{Contradiction:} \textit{SUBJ ADV won't need to check it again}.

    \item
        \textbf{Premise:} \textit{SUBJ can ADV find bargains in Tunis}.\\
        \textbf{Entailment:} \textit{SUBJ will ADV find good deals in Tunis}.\\
        \textbf{Contradiction:} \textit{SUBJ will ADV discover that everything is expensive in Tunis}.

    \item
        \textbf{Premise:} \textit{His friend, SUBJ, is ADV a foreigner}.\\
        \textbf{Entailment:} \textit{SUBJ ADV is from another country}.\\
        \textbf{Contradiction:} \textit{SUBJ ADV is a native here}.
\end{enumerate}

\subsection{Unattributed-information adverbs}

\paragraph{Adverbs:} \textit{reportedly}, \textit{allegedly}, \textit{supposedly}, \textit{apparently}, \textit{ostensibly}.

\begin{enumerate}
    \item
        \textbf{Premise:} \textit{The German government ADV opposed the quotas}.\\
        \textbf{Entailments:} \textit{The German government ADV was against the quotas};
            \textit{The German government may have supported the quotas}.\\
        \textbf{Contradiction:} \textit{The German ADV supported more quotas}.

    \item
        \textbf{Premise:} \textit{The celebration had been postponed, ADV because of the Gulf War}.\\
        \textbf{Entailments:} \textit{Someone said that the celebration was postponed because of the Gulf War}; \textit{The Gulf War may have had no effect on the celebration}.\\
        \textbf{Contradiction:} \textit{The Gulf War ADV had no effect on the celebration}.
\end{enumerate}

\subsection{Degree Adverbs}

\paragraph{Adverbs:} \textit{at least}, \textit{at a minimum}, \textit{nearly}, \textit{approximately}.

\begin{enumerate}
    \item
        \textbf{Premise:} \textit{Lantau covers ADV twice the area of Hong Kong Island}.\\
        \textbf{Entailment:} \textit{Lantau is at least as big as Hong Kong Island}.\\
        \textbf{Contradiction:} \textit{Hong Kong Island is at least as big as Lantau}.

    \item
        \textbf{Premise:} \textit{At the moment ADV 140 persons are working to curtail the fire}.\\
        \textbf{Entailment:} \textit{Many people are fighting the fire}.\\
        \textbf{Contradiction:} \textit{Nobody is fighting the fire}.
\end{enumerate}

\subsection{Obviousness Adverbs}

\paragraph{Adverbs:} \textit{blatantly}, \textit{obviously}, \textit{clearly}, \textit{ostentatiously}, \textit{noticeably}, \textit{visibly}, \textit{conspicuously}.

\begin{enumerate}
    \item
        \textbf{Premise:} \textit{Castro ADV backed the rebels}.\\
        \textbf{Entailments:} \textit{Castro helped the rebels};
            \textit{Castro tried to help the rebels}.\\
        \textbf{Contradiction:} \textit{Castro tried to stop the rebels}.

    \item
        \textbf{Premise:} \textit{The students were ADV drunk}.\\
        \textbf{Entailment:} \textit{The students were surely drinking too much}.\\
        \textbf{Contradiction:} \textit{The students renounced alcohol}.
\end{enumerate}

\section{Details on the Synthetic Dataset}
\label{sec:appendix-synthetic}

\subsection{Fillers for the human-subject slot}

\textit{James}, \textit{Mary}, \textit{Robert}, \textit{Patricia}, \textit{John}, \textit{Jennifer}, \textit{Michael}, \textit{Linda}, \textit{David}, \textit{Elizabeth}, \textit{William}, \textit{Barbara}, \textit{Richard}, \textit{Susan}, \textit{Joseph}, \textit{Jessica}, \textit{Thomas}, \textit{Sarah}, \textit{Charles}, \textit{Karen}, \textit{Li}, \textit{Wei}, \textit{Fang}, \textit{Xiuying}, \textit{Na}, \textit{Priya}, \textit{Rahul}, \textit{Divya}, \textit{Abhishek}, \textit{Ishita}, \textit{Melokuhle}, \textit{Omphile}, \textit{Iminathi}, \textit{Lisakhanya}, \textit{Lethabo}, \textit{Ivaana}, \textit{Malik}, \textit{Pipaluk}, \textit{Aputsiaq}, \textit{Nivi}.

\subsection{Unattributed-information adverbs}

\paragraph{Adverbs:} \textit{reportedly}, \textit{allegedly}, \textit{supposedly}, \textit{apparently}, \textit{ostensibly}.

\paragraph{Actions:} \textit{the wedding}, \textit{the marriage}, \textit{buying the house}, \textit{selling the car}, \textit{moving away}, \textit{staying in Canberra}, \textit{delaying the funeral}, \textit{the arrangement}, \textit{the lawsuit}.

\paragraph{Premise:} \textit{SUBJ1 said that SUBJ2 ADV opposed ACTION.}

\paragraph{Entailments:} 

\begin{enumerate}
    \setlength\itemsep{0em}
    \item \textit{SUBJ1 said that SUBJ2 may have opposed ACTION.}
    \item \textit{SUBJ1 is not sure that SUBJ2 opposed ACTION.}
\end{enumerate}

\paragraph{Contradictions:}

\begin{enumerate}
    \setlength\itemsep{0em}
    \item \textit{SUBJ1 is sure that SUBJ2 opposed ACTION.}
    \item \textit{SUBJ1 is sure that SUBJ2 did not support ACTION.}
\end{enumerate}

\subsection{Degree adverbs}

\paragraph{Adverbs:} \textit{at least}, \textit{at a minimum}, \textit{nearly}, \textit{approximately}.

\subsubsection{Properties of people}

\paragraph{Properties:} \textit{net worth}, \textit{knowledge}, \textit{manners}, \textit{fan base}, 
\textit{culpability}.\footnote{Unlike in case with adverbs and subject-slot fillers, where all combinations are used,
properties and adjectives in this and the next subclass are used in parallel. I.e., when the \textit{i}'th adjective 
from the first list is used in the premise, the corresponding \textit{i}'th property and adjective from the second 
list will be used in the hypotheses.}

\paragraph{Adjectives:}
\begin{itemize}
    \setlength\itemsep{0em}
    \item \textbf{Adjective 1}: \textit{rich}, \textit{erudite}, \textit{polite}, \textit{popular}, \textit{guilty}.
    \item \textbf{Adjective 2}: \textit{big}, \textit{extensive}, \textit{good}, \textit{large}, \textit{high}.
\end{itemize}

\paragraph{Premise:} \textit{SUBJ1 is ADV twice as ADJ1 as SUBJ2.}
\paragraph{Entailment:} \textit{SUBJ1's PROPERTY is/are at least as ADJ2 as SUBJ2's.}
\paragraph{Contradiction:} \textit{SUBJ2's PROPERTY is/are at least as ADJ2 as SUBJ1's.}

\subsubsection{Properties of objects}

\paragraph{Subjects:} \textit{the truck}, \textit{the house}, \textit{the hotel}, \textit{the ship}, \textit{the wagon}, \textit{the car}, \textit{the tree}.

\paragraph{Properties:} \textit{age}, \textit{weight}, \textit{height}, \textit{width}, \textit{price}.

\paragraph{Adjectives:}
\begin{itemize}
    \setlength\itemsep{0em}
    \item \textbf{Adjective 1}: \textit{old}, \textit{heavy}, \textit{tall}, \textit{wide}, \textit{expensive}.
    \item \textbf{Adjective 2}: \textit{great}, \textit{big}, \textit{big}, \textit{big}, \textit{high}.
\end{itemize}

\paragraph{Premise:} \textit{SUBJ1 is ADV twice as ADJ1 as SUBJ2.}
\paragraph{Entailment:} \textit{The PROPERTY of SUBJ1 is at least as ADJ2 as that of SUBJ2.}
\paragraph{Contradiction:} \textit{The PROPERTY of SUBJ2 is at least as ADJ2 as that of SUBJ1.}

\subsubsection{Quantities}

\paragraph{Times:} \textit{at the moment}, \textit{now}, \textit{these days}, \textit{this month}, 
\textit{this week}.\footnote{Similarly to the two previous subclasses,
times, numbers, activities, and related-person groups in this subclass are used in parallel. 
I.e., when the \textit{i}'th time, number, related-person group, and activity are used in the premise, 
the corresponding \textit{i}'th activity will be used in the hypotheses.}

\paragraph{Numbers:} \textit{two dozen}, \textit{thirty}, \textit{fifty}, \textit{140}.

\paragraph{Related-person groups:} \textit{friends}, \textit{relatives}, \textit{acquaintances}, \textit{coworkers}.

\paragraph{Activities:} \textit{working on this}, \textit{helping with the move}, \textit{coming to visit us}.

\paragraph{Premise:} \textit{TIME ADV NUMBER of SUBJ's RELATED\_PERSONS are ACTIVITY.}
\paragraph{Entailment:} \textit{Many people are ACTIVITY.}
\paragraph{Contradiction:} \textit{Nobody is ACTIVITY.}

\subsection{Obviousness adverbs}

\paragraph{Adverbs:} \textit{blatantly}, \textit{obviously}, \textit{clearly}, \textit{ostentatiously}, \textit{noticeably}, \textit{visibly}, \textit{conspicuously}.

\paragraph{Actions:}\footnote{Similarly to adjectives and properties in the case of degree adverbs above, actions of different types are used in parallel. I.e., when the \textit{i}'th element from the first list is used in the premise, corresponding \textit{i}'th elements from other lists will be used in the hypotheses.}

\begin{itemize}
    \setlength\itemsep{0em}
    \item \textbf{Action 1}: \textit{backed}, \textit{supported}, \textit{criticized}, \textit{provoked}, \textit{brainwashed}.
    \item \textbf{Action 2, past}: \textit{helped}, \textit{encouraged}, \textit{disparaged}, \textit{incited}, \textit{indoctrinated}.
    \item \textbf{Action 2, infinitive}: \textit{help}, \textit{encourage}, \textit{disparage}, \textit{incite}, \textit{indoctrinate}.
    \item \textbf{Action 3, past}: \textit{stopped}, \textit{deterred}, \textit{praised}, \textit{calmed}, \textit{deprogrammed}.
    \item \textbf{Action 3, infinitive}: \textit{stop}, \textit{deter}, \textit{praise}, \textit{calm}, \textit{deprogram}.
\end{itemize}

\paragraph{Premise:} \textit{SUBJ1 ADV ACTION1 SUBJ2.}

\paragraph{Entailments:}
\begin{enumerate}
    \setlength\itemsep{0em}
    \item \textit{SUBJ1 ACTION2\_PAST SUBJ2.}
    \item \textit{SUBJ1 tried to ACTION2\_INF SUBJ2.}
\end{enumerate}

\paragraph{Contradictions:}
\begin{enumerate}
    \setlength\itemsep{0em}
    \item \textit{SUBJ1 ACTION3\_PAST SUBJ2.}
    \item \textit{SUBJ1 tried to ACTION3\_INF SUBJ2.}
\end{enumerate}




\end{document}